\documentclass[letterpaper, 10 pt, conference]{ieeeconf}  

\IEEEoverridecommandlockouts                              

\usepackage[T1]{fontenc}    
\usepackage{hyperref}       
\usepackage{url}            
\usepackage{booktabs}       
\usepackage{amsfonts}       
\usepackage{nicefrac}       
\usepackage{microtype}      
\usepackage{amsmath}                      
\usepackage{mathtools}
\usepackage{graphicx}
\usepackage{float}
\usepackage{subcaption}
\usepackage{wrapfig}
\usepackage{verbatim}
\usepackage{xcolor}
\usepackage{verbatim}
\usepackage{amssymb}
\usepackage{mathtools, nicefrac}
\usepackage{flushend}

\title{Vid2Param: Modelling of Dynamics Parameters from
  Video}
\author{Martin Asenov$^{1}$, Michael Burke$^{1}$, Daniel Angelov$^{1}$, Todor Davchev$^{1}$, Kartic Subr$^{1}$, Subramanian Ramamoorthy$^{1}$
\thanks{M. Asenov and D. Angelov are supported by the Engineering and Physical
Sciences Research Council (EPSRC), as part of the CDT in RAS at Heriot-Watt
University and The University of Edinburgh. K. Subr is supported by a
Royal Society University Research Fellowship. M. Burke and S. Ramamoorthy acknowledge funding from the Alan Turing Institute for the project, Safe AI for Surgical 
Assistance.}
\thanks{$^{1}$Institute of Perception, Action and Behaviour (IPAB), School of Informatics, The University of Edinburgh, EH8 9AB, UK
(email: {\tt\small m.asenov@ed.ac.uk})}%
}%

\graphicspath{{figures2/}}

 \newcommand{\ignore}[1] {#1} 

\newcommand{\Martin}[1] {\ignore{\textcolor{red}{MA: \ignore{#1}}}}

\newcommand{\cmmnt}[1]{\ignorespaces}

\newcommand{\eq}[1]{

\begin{equation}
  {#1}
\end{equation}
}

\newcommand\offsetl{-8mm}
\newcommand\offset{-4mm}
\newcommand\offsets{-7mm}

\begin{document}

\maketitle
\thispagestyle{empty}
\pagestyle{empty}

\begin{abstract}

Videos provide a rich source of information, but it is generally hard to extract dynamical parameters of interest. Inferring those parameters from a video stream would be beneficial for physical reasoning. Robots performing tasks in dynamic environments would benefit greatly from understanding the underlying environment motion, in order to make future predictions and to synthesize effective control policies that use this inductive bias. {\textit{Online}} physical reasoning is therefore a fundamental requirement for robust autonomous agents. When the dynamics involves multiple modes (due to contacts or interactions between objects) and sensing must proceed directly from a rich sensory stream such as video, then traditional methods for system identification may not be well suited. We propose an approach wherein fast parameter estimation can be achieved directly from video. We integrate a physically based dynamics model with a recurrent variational autoencoder, by introducing an additional loss to enforce desired constraints. The model, which we call Vid2Param, can be trained entirely in simulation, in an end-to-end manner with domain randomization, to perform online system identification, and make probabilistic forward predictions of parameters of interest. This enables the resulting model to encode parameters such as position, velocity, restitution, air drag and other
physical properties of the system.  We illustrate the utility of this in physical experiments wherein a PR2 robot with a velocity constrained arm must intercept an unknown bouncing ball with partly occluded vision, by estimating the physical parameters of this ball directly from
the video trace after the ball is released.

\end{abstract}

\section{Introduction}

There is an ever growing need to perform robotic tasks in unknown
environments. Reasoning about observed dynamics using ubiquitous sensors such as
video is therefore highly desirable for practical robotics. Traditionally, the complexity of this
reasoning has been avoided by investing in fast actuators
\cite{furukawa2006dynamic} \cite{namiki2003development} and using very accurate
sensing  \cite{kapur2005gesture}. In emerging field applications of
robotics, the reliance on such infrastructure may need to be decreased \cite{hastie2018orca}, while
the complexity of tasks and environment uncertainty has increased \cite{erickson2018deep}. As such,
there is a need for better physical scene understanding from low-cost sensors and the ability to make
forward predictions of the scene, so as to enable planning and control.

Recent advances have enabled video prediction conditioned on observations
\cite{lee2018stochastic} and reasoning about complex physical phenomena
\cite{takahashi2019video}. Video streams provide a rich source of information, but
it is often challenging to acquire the compressed structured representations of interests. Techniques for system
identification, originally developed for process control domains, are aimed
at this problem \cite{ljung2001system}.  There are a number of different
approaches to estimating parameters \cite{johansen1995identification}, and
sometimes even model structure \cite{li2016identification}, from observed data.

Acquiring reduced representations of the environment and performing system
identification have historically been disjointly solved, despite the rich
contextual information images often contain. This may lead to slower inference
or even failure of the optimization should the state space model reduction
be inaccurate. Yet, solving the problem jointly brings a set of practical
challenges. First, it is difficult to acquire the necessary training data to
cover the possible variations of the task of interest and generalize to unseen
data. Secondly, a model needs to learn to perform probabilistic inference of
parameters of interest and generation from a sequence of images, and capture long-term dependencies. From a practical robotics perspective, the
model needs to be sufficiently fast to be able to perform system identification
on-the-fly, while performing inference directly from images.

In this paper, we focus on the case where a robot must perform robust system
identification {\textit{online}}, directly from a rich sensory stream such as
video (including the implicit tasks of detecting and tracking objects).
We pose the problem as learning an end-to-end model, where we regress from
videos directly to parameters of interest. Furthermore, we structure the
learned model so as to be able to make probabilistic future predictions in the
latent space, to enable appropriate action selection. This allows for
interacting in relatively unknown environments when system identification must
be performed on-the-fly for the successful completion of a task.

\begin{figure*}[htbp]
\centering
\vspace{1mm}
\includegraphics[width=0.85\linewidth]{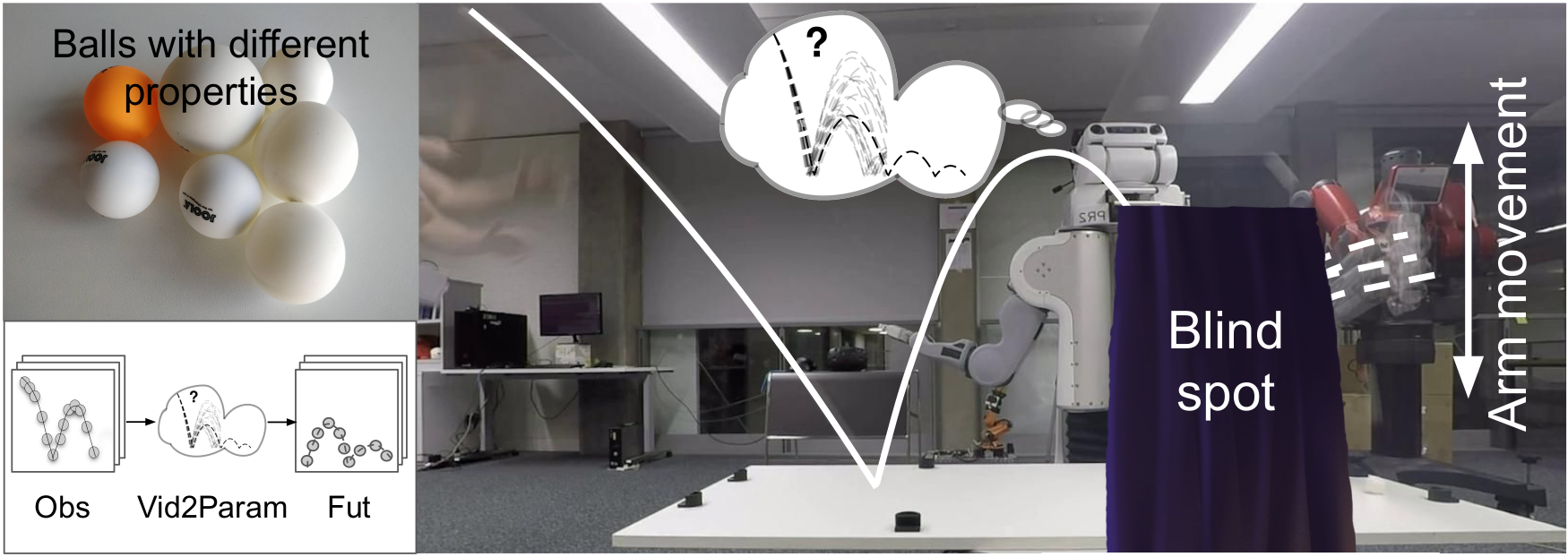}
\vspace{-2mm}
\caption{\textbf{Overview and experimental setup.} 
 We are interested in reasoning about the dynamics in an environment, by using a single video
stream as a sensory input. To demonstrate the utility of our approach, we use a slowly actuated
robotics arm to perform a 'stopping a bouncing ball' experiment, where the
physical properties of the ball or height of the table are not known a priori and occlusions are
present. Our model is able to perform online inference of the parameters of interests
and generate plausible future trajectories. 
(Please refer to
   supplementary video for additional results.)}
\label{robotexpfig}
\vspace{\offsets}
\end{figure*}

We present a model that enforces physical conformity between videos and dynamical
parameters of interests. We integrate an analytic simulator with a recurrent
latent model \cite{chung2015recurrent} by introducing an additional loss term for encoder-decoder
mapping from a given sensory input (vision) to physical parameters and states
(position, velocity, restitution factor, gravity, etc. in a parametric
description of a physics-based model). We show that such a model can be trained
with suitable domain randomization \cite{tobin2017domain} in simulation and deployed in a real
physical system. This model, which we call Vid2Param, allows for forward
predictions envisioning possible future trajectories based on uncertainty in the
estimate of the physical parameters. We demonstrate that such a model can indeed
perform accurate system identification directly from videos by demonstrating that we achieve similar levels of performance as traditional system identification methods,
which have access to ground truth starting trajectories. We perform experiments on simulated and real recorded videos of a bouncing ball with different physical properties. To illustrate the utility of this capability, we demonstrate this model on the task of intercepting a bouncing ball with a relatively slow moving robot arm and standard visual sensing.

\section{Related Work}

\subsection{System Identification}


System identification (SysID) is concerned with
the problem of determining the structure and parameters of a dynamical system, for subsequent use in controller design. The best developed versions of system identification methods focus on the case of linear time-invariant (LTI) systems, although almost all of these methods have also been extended to the case of nonlinear and hybrid dynamical systems. With these more complex model structures, the computational complexity of identification can be relatively high even for moderately sized data sets.
Examples of system identification procedures that could be applied to our problem domain, including the additional step of reducing model order, include the Eigen system realization algorithm \cite{juang1985eigensystem} and Balanced POD (BPOD)
\cite{rowley2005model} (which theoretically obtain the same reduced models
\cite{ma2011reduced}), and the use of feedforward neural networks \cite{chen1990non}. BPOD can be
viewed as an approximation to another popular method, Balanced truncation
(BT) \cite{safonov1989schur}, which scales to larger systems. 

Another way to approach the problem of identification is frequency domain decomposition \cite{brincker2001modal}. Recent approaches in this vein include DMD
\cite{kutz2016dynamic} and Sindy \cite{brunton2016discovering}, which allow for
data driven, model-free system identification and can scale to high-dimensional
data. When performing SysID directly from a rich sensory stream like video, it is not always clear what the optimal reduced representation should be \cite{achille2018separation}. We exploit the fact that a physics based model of objects can provide useful regularisation to an otherwise ill-posed identification problem.

\subsection{Simulation alignment}

When a parametric system model is available, simulation alignment can be
performed to identify the parameters of the system. A standard approach is to perform least squares minimization or maximum likelihood estimation, for instance computing best fit parameters to align simulator traces to observed data \cite{wu2015galileo}. When simulation calls are expensive, a prior over the parameter space can be enforced, e.g. Gaussian Processes, and Bayesian Optimization can be used \cite{romeres2016line} \cite{ramos2019bayessim} \cite{lopez2017adaptable} \cite{asenov2019active}. Our approach is closely related to \cite{zhang2016learning} as we use supervision
during the training phase of our model, and then use this learned approximation at test time. We also employ domain randomization while training our model \cite{peng2018sim} and our work follows a similar line of reasoning to that of \cite{chebotar2018closing}, which aims to align a simulator to real world observations as the model is being trained. We focus on the problem of aligning a model to online observations at test time, for predictive purposes.


\subsection{Learnable Physics Engines}

There has been increasing interest in learnable physics engines - for example learning complex factorization (at the object or particle level) from data  \cite{mrowca2018flexible} \cite{battaglia2016interaction}, using particle based networks for fluid dynamics \cite{schenck2018spnets} and in robotics \cite{degrave2019differentiable}. By representing the problem in terms of a set of learnable components (graph representing objects/particle and relations, Navier Stokes equations, linear complementary problem for the above mentioned tasks) a physics engine can be learned from raw data. Similar approaches have  been shown to scale to video data \cite{watters2017visual}. We explore the complementary problem of system identification (with an analytical or learned simulator), and propose a direct optimization approach by learning an inverse probabilistic physics engine. This builds upon ideas presented in \cite{wu2015galileo}, where an analytical simulator is used with traditional system identification approaches. Closely related work is presented in \cite{purushwalkam2019bounce}, where surface properties are learned using Physics and Visual Inference Modules.

A related question to learning interactions between objects is that of learning a state space system to represent these. This has been explored for individual objects \cite{karl2016deep} \cite{fraccaro2017disentangled}, by using Kalman and Bayes filters for learning. State models and predictions have recently been explored in the context of videos involving multiple objects \cite{hsieh2018learning} through the use of Spatial Transformer Networks \cite{jaderberg2015spatial} and decompositional structures for the dynamics, as well as integrating the differential equations directly into a network \cite{jaques2019physics}.

\subsection{Variational Autoencoder}

Variational Autoencoders (VAEs) \cite{kingma2013auto} have been extensively
applied to image and video generation \cite{eslami2016attend} \cite{hsieh2018learning}. Recently, VAEs have been used in reinforcement learning to improve generalization by learning a master policy from a set of similar MDPs \cite{arnekvist2018vpe}. Closely related work is that of \cite{ajay2018augmenting} where Variational RNNs are used to learn `residual physics' \cite{zeng2019tossingbot} \cite{shi2018neural}. The addition of loss terms to the reconstruction and KL error terms have also been proposed, allowing for enforcement of multiple desired constraints \cite{pmlr-v87-hristov18a} \cite{Angelov:2019:UCA:3306127.3331841}. We extend this line of work, by demonstrating that such constraints can be applied in a recurrent model to satisfy physics properties.











\begin{figure*}[htbp]
\centering
\vspace{2mm}
\includegraphics[width=0.78\linewidth]{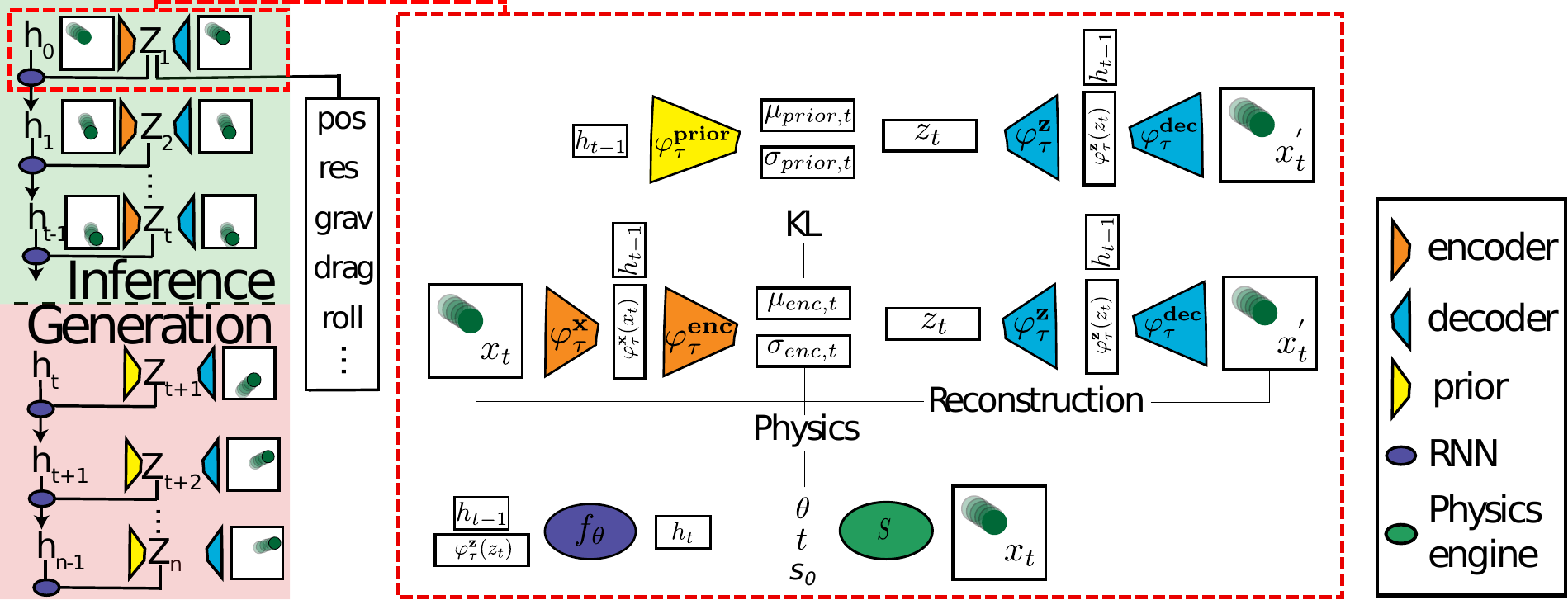}
\vspace{-2mm}
\caption[Overview]{\textbf{Technical details and notation.} We propose an end-to-end model for performing system identification directly from unstructured input, such as video. We based our model on a Variational Recurrent Neural Network (VRNN) \cite{chung2015recurrent}. To encode the physical properties of interests we propose an additional Gaussian negative log likelihood loss between the parameters of interest and part of the latent space. The inference and generation overview of the model (left) and the training procedure at each frame (right) can be seen in the figure.}
\label{overview}
\vspace{\offsetl}
\end{figure*}

\section{Vid2Param for online system identification from videos}

\textbf{Problem formulation}
Given a set of sensory observations $x_{1:t-1}$, we are interested in predicting future observations
$x_{t}$ using a low-dimensional dynamics representation $p\left(z_{t}|z_{1:t-1}\right)$.
\begin{equation}
\begin{multlined}
 p\left(x_{t}|x_{1:t-1}\right)
  =  \int p\left(x_{t}|z_{t}\right) p\left(z_{t}|z_{1:t-1}\right)p\left(z_{1:t-1}|x_{1:t-1}\right)dz\\
  \hspace*{0.5cm} = \int p\left(x_{t}|z_{t}^{\prime},\theta_{t}\right) p\left(z_{t}^{\prime},\theta_{t}|z_{1:t-1}^{\prime},\theta_{1:t-1}\right)\\
  p\left(z_{1:t-1}^{\prime},\theta_{1:t-1}|x_{1:t-1}\right)dz\hspace*{1cm}
\end{multlined}
\label{factorization}
\end{equation}
Here, we decompose the latent space $z=[z^{\prime},\theta]$ into the physical dynamics parameters of interest,
$\theta$, and a remaining $z^{\prime}$ term, used for image reconstruction and to capture potentially un-modelled effects.
We illustrate how this model can be learned in an end-to-end fashion from videos, and how we can use the
predictions in the latent space for model predictive control. 

\textbf{Variational Recurrent Neural Networks}
We implement the inference process above using a modified recurrent VAE (VRNN)
 \cite{chung2015recurrent}. A VRNN consists of an RNN encoding the dynamics
 of the sequence, and a VAE conditioned on those dynamics, by including
 the hidden state of the RNN at each step. As shown in fig.\ref{overview}, the variational auto encoder
$\varphi_{\tau}^{enc}$ ($\tau$ denotes the trainable parameters of the neural network) takes in
a latent representation of the input $\varphi^{x}_{\tau}(x_{t})$, in addition to the
 hidden state of an RNN, $h_{t-1}$. The encoder network then produces the mean and variance components, $\mu_{enc,t}$ and $\sigma_{enc,t}$ respectively, of a multivariate Gaussian distribution, $q$, that are also conditioned on $h_{t-1}$, thereby
 capturing information about the dynamics of the sequence until $t$: 
\begin{equation}
\begin{multlined}
   q\left(z_{t}|x_{\leq t},z_{<t}\right)  =
   \mathcal{N}\left(\mu_{enc,t},diag\left({\sigma_{enc,t}^{2}}\right)\right)\text{,}\\ 
 \text{where }   \mu_{enc,t},\sigma_{enc,t} = \varphi^{enc}_{\tau}\left(\varphi^{x}_{\tau}\left(x_{t}\right),h_{t-1}\right)
\end{multlined}
\end{equation}

Similarly, the generative component of the VAE is also expanded by including the hidden state $h_{t-1}$:
\begin{equation}
\begin{multlined}
   p\left(x_{t}|z_{\leq t},x_{<t}\right) =\mathcal{N}\left(\mu_{dec,t},diag\left({\sigma_{dec,t}^{2}}\right)\right)\text{,} \\
 \text{where }  \mu_{dec,t},\sigma_{dec,t}=\varphi^{dec}_{\tau}\left(\varphi^{z}_{\tau}\left(z_{t}\right),h_{t-1}\right)
\end{multlined}
\end{equation}

This means that the prior is no longer a standard normal
distribution $\mathcal{N}\left(0, 1\right)$ as in the case of a vanilla VAE,
but is specified by a network conditioned on the hidden state $h_{t-1}$:
\begin{equation}
\begin{multlined}
   p\left(z_{t}|x_{<t},z_{<t}\right) =\mathcal{N}\left(\mu_{prior,t},diag\left(\sigma_{prior,t}^{2}\right)\right)\text{,} \\
 \text{where } \mu_{prior,t},\sigma_{prior,t} =\varphi^{prior}_{\tau}\left(h_{t-1}\right)
\end{multlined}
\end{equation}
 
Finally, the RNN updates its hidden state $h_{t}$ with a transition function
$f_{\psi}$ (with parameters $\psi$) by taking in a latent representation of
the input $\varphi^{x}_{\tau}(x_{t})$ a sample from
$\varphi^{z}_{\tau}(z_{t})$, and the previous hidden state $h_{t-1}$:

\eq{h_{t}=f_{\psi}\left(\varphi^{x}_{\tau}\left(x_{t}\right),\varphi^{z}_{\tau}\left(z_{t}\right),h_{t-1}\right)}
VAEs are trained by maximizing an evidence lower bound \cite{kingma2013auto}. Given the VRNN modifications, the overall loss, including a Kullback-Leibler (KL) divergence and reconstruction loss term, becomes:
\begin{equation}
\begin{multlined}
  \mathbb{E}_{q(z\leq T|x\leq T)}\Bigl[\sum_{t=1}^{T} 
  \log p\left(x_{t}|z_{\leq t},x_{<t}\right) -\\
  \mathrm{KL}\left(q\left(z_{t}|x_{\leq t},z_{<t}\right)\|p\left(z_{t}|x_{<t},z_{<t}\right)\right)\Bigr].
\end{multlined}
\end{equation}
In summary, the VRNN is a modified VAE that makes use of a learned low-dimensional dynamics model to make sequential predictions. However, the VRNN provides no guarantees that the latent representation is meaningful, and as a result is unsuited for use in  control or for system identification.

\textbf{Vid2Param}
We propose combining the encoder-decoder factorization with dynamics modelling
in the latent space $z$, conditioned on parameters of interest, $\theta$. We introduce an additional loss to the standard VRNN to encourage encoding of
 physically meaningful parameters, by including a Gaussian negative log
 likelihood \cite{nix1994estimating} loss between part of the latent space $z$ and the physical parameters $\theta$ we are
 interested in (e.g., gravity, restitution, position, etc., in the case of a
 bouncing ball). The loss terms are scaled with non-negative numbers $\alpha$, $\beta$
 and $\gamma$ and we let $N_\theta$ represent the size
 of the parameter vector.
\begin{equation}
\begin{multlined}
  \mathbb{E}_{q(z\leq T|x\leq T)}\Bigl[\sum_{t=1}^{T}\Big(\alpha\log p\left(x_{t}|z_{\leq t},x_{<t}\right)- \\
  \mathrm{\beta KL}\left(q\left(z_{t}|x_{\leq t},z_{<t}\right)\|p\left(z_{t}|x_{<t},z_{<t}\right)\right) + \\
  \gamma\sum_{i=1}^{N_\theta}\big(\frac{1}{2} \ln (2 \pi)+\frac{1}{2} \ln \left((\sigma_{enc,t}^{i})^{2}\right)+ \frac{(\theta^{i}_{t} -\mu_{enc,t}^{i})^{2}}{2 (\sigma_{enc,t}^{i})^{2}}\big)\Big)   \Bigr]
\end{multlined}
\label{myloss}
\end{equation}

Given trained networks, we can now perform probabilistic inference of physical parameters from sensory
data such as a sequence of images using the factored portion of the latent space directly.
\eq{P(\theta_{t}^{i}|x_{\leq t})=\mathcal{N}(\mu_{enc,t}^{i},
  \sigma_{enc,t}^{i})}

Additionally, we can sample plausible future extrinsic properties (eg. positions) by recursively updating the model predictions to generate future states.
\begin{equation}
\begin{multlined}
  P(\theta_{t}^{i}|x_{< t},\theta_{< t})=   \mathcal{N}(\mu_{prior,t}^{i},  \sigma_{prior,t}^{i})
\end{multlined}
\end{equation}

As a final modification, we exclude $x_{t}$ from the recurrent step, since all necessary information is already present in $z_{t}$. This speeds up the prediction in the latent space, as $x$ does not need to be reconstructed and fed back into the network at every step. As such we can make recursive future predictions entirely in the latent space,
\eq{h_{t}=f_{\psi}\left(\varphi^{z}_{\tau}\left(z_{t}\right),h_{t-1}\right).}


To summarise, the contributions of this paper include:
\begin{enumerate}
\item Extension of the VRNN model with a loss term to encode dynamical properties.

\item Enabling faster future predictions in the latent space, along with uncertainty quantification through the identified parameters.

\item Evaluation of speed and accuracy of identification, against alternate approaches to system identification

\item Demonstration on a physical robotic system, in a task requiring interception of a bouncing ball whose specific physical parameters are unknown a priori, requiring online identification from the video stream.
\end{enumerate}








\section{Experiments}
First, we perform a series of experiments on simulated videos, when ground truth is explicitly available. We compare our method against existing system identification methods, evaluating speed of estimation and accuracy of the identified parameters. Next, we evaluate our method on a set of real videos. Finally, we perform a physical experiment involving online system identification from a camera feed. 
\subsection{Setup}
\label{ref_setup}

We use a bouncing ball as an example hybrid dynamical system. This is a particularly useful example, as the dynamics of the ball vary depending on the ball state, making system identification particularly challenging from high dimensional sensor data using classical techniques. The governing dynamics of the bouncing ball can be described using the following set of ordinary differential equations: 
\begin{equation}\label{bouncing_ball}
S=\begin{cases}
s_{t}=s_{0}+\dot{s}_{0}t+\cfrac{1}{2}\ddot{s}t^{2},\ddot{s}=g-d & \text{Free fall}\\
\dot{s}_{t}^{y}=-e\dot{s}_{t-1}^{y},\text{when }\dot{s}_{t}^{y}<0; s_{t}^{y}=0 & \text{Bounce}\\
\dot{s}_{t}^{x}=r\dot{s}_{t-1}^{x},\text{when }\ddot{s}_{t}^{y}=0 & \text{Rolling}
\end{cases}
\end{equation}

We use $s,\dot{s},\ddot{s}$ for the current position, velocity and acceleration respectively, $e$ is the coefficient of restitution and $r$ the rolling resistance. Additional dynamic effects are often observed such as air drag $d=-c\dot{s}\sqrt{(\dot{s}^{x})^{2}+(\dot{s}^{y})^{2}}/m$,
 where $c$ is the drag constant and $m$ is mass. Thus the acceleration becomes
 $\ddot{s}=g+d$, where $g$ is the gravitational force  ($g^{x}=0$). As such, the system is completely determined by the initial state
 of the system $s,\dot{s},\ddot{s}$ and its physical properties
 $e,r,m,c,g\in\theta$. It should be noted that the real world behaviour of any specific ball could deviate from this model depending on its shape, initial spin (Magnus effect), the presence of wind, and so on.

 We make the following assumptions: 1) there is a single moving object, the bouncing ball 2) the ball
 bounces off a flat surface. We do not assume to know the physical properties of the ball, the height of the surface or the exact velocities of the ball,  and we use a single low quality camera for sensing.


 We use a parallel adaptive ODE solver to simulate 
 data described by eq.\ref{bouncing_ball}. We use these simulated trajectories to
 generate a sequence of images. We generate 10000 training and 100 test videos,
 with 200 timesteps/10 seconds, $28 \times 28$, with $e\in[0.6,1.0]$,
 $g\in[-6.81,-12.81], d\in[0.05,0.0005], r\in[0.0,0.7]$. We split the parameters
 into 10 sub-ranges and alternately use them for training and testing, so no parameters used in the training data are available in the test data.
 
 For the real videos and robot experiment, we trained a separate model with 5000 videos, $100 \times 50$
 with 75 timesteps/10 seconds and the same physical parameters. Additionally, we add motion blur based on the velocity and black-out part of the frames to
 account for some of the missing/noisy data typically exhibited when using
 low-cost cameras. Additionally, we randomize the height of the plane on which
 the ball bounces. Our encoder-decoder network follows a similar architecture to
 \cite{chen2016infogan} and for the RNN we use a standard LSTM network. We set
 $\alpha=1$, $\beta=1$ and $\gamma=10$ throughout our experiments
 (eq.\ref{myloss}). We use an NVidia 1080 Ti for training and laptop NVidia
 Quadro M2000M GPUs for testing the model. We use MSE as an accuracy metric throughout the paper and normalize positions and parameters.
 
 \subsection{System identification}
\label{SysIDexp}
In this experiment, we evaluate the speed and accuracy of the proposed method against different simulation alignment approaches. We evaluate the likelihood of the observed trajectory with respect to simulated trajectories by performing MLE estimation (similarly to \cite{wu2015galileo}). We sample 2000 trajectories by placing a uniform prior over the physical parameters and compare the simulated trajectories against those observed. We select the parameters which generated the least error between the trajectories, and compare these against parameters that generated the observed trajectory.

Secondly, we extend this by imposing a Gaussian Process prior over the
parameters of interest and performing Bayesian Optimization
\cite{gonzalez2016gpyopt} \cite{ramos2019bayessim}. We use Expected Improvement as an acquisition function and use 20 optimisation steps. The baselines have access to
the initial velocity, $n$ number of positions (as such we don't need a tracker
as in \cite{wu2015galileo}) and an optimized ODE solver for sampling.

\begin{figure}[htbp]
\centering
\begin{subfigure}[b]{0.66\linewidth}
  \centering
  \includegraphics[width=1.0\linewidth]{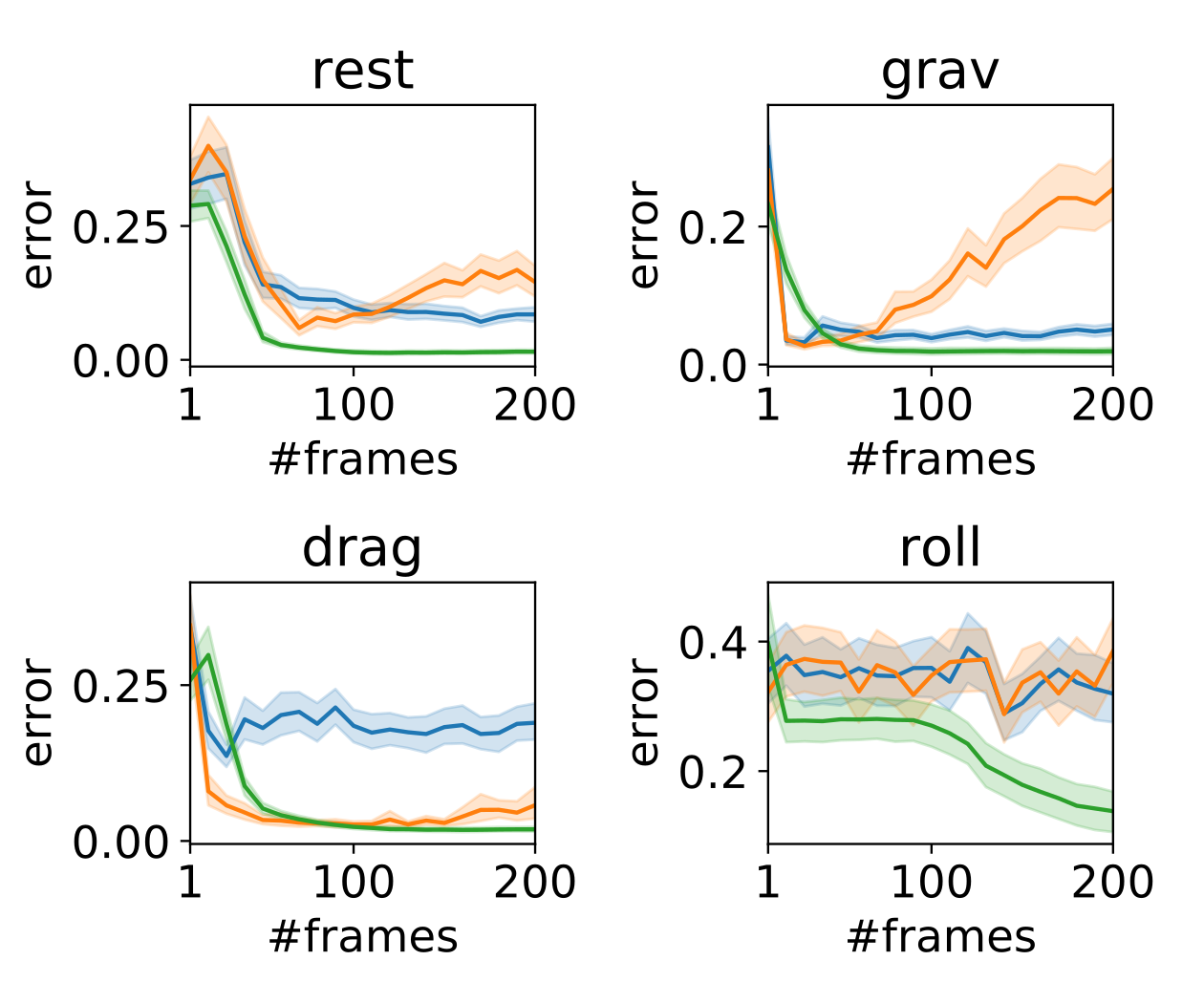}
  \caption{Accuracy}
\end{subfigure}%
\begin{subfigure}[b]{0.33\linewidth}
  \centering
  \includegraphics[width=1.0\linewidth]{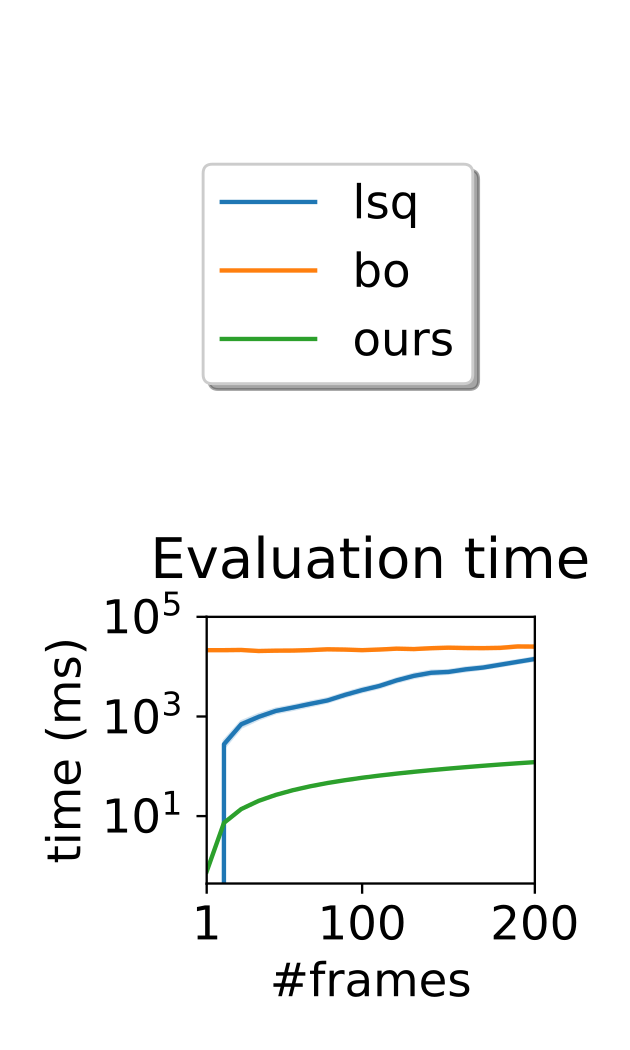}
  \caption{Speed}
\end{subfigure}
\caption[sysID]{\textbf{Performance of different system identification methods with variable number of observed frames.} (a) Overall error of the predicted normalized parameters (b) Speed of computation. We denote \cite{wu2015galileo} as 'lsq' and \cite{gonzalez2016gpyopt} as 'bo'.}
\label{sysID}
\vspace{\offset}
\end{figure}

In contrast, our trained model receives {\textit{only}} the video as an input, and no other parameters.  Speed and accuracy benchmarks are shown in fig.\ref{sysID}. It can be seen that our method has similar or better performance, despite not having access to the initial ground truth trajectory  of the ball.  

\subsection{Forward predictions}

\begin{figure}[htbp]
\centering
\begin{subfigure}[b]{0.90\linewidth}
  \centering
  \includegraphics[width=1.0\linewidth]{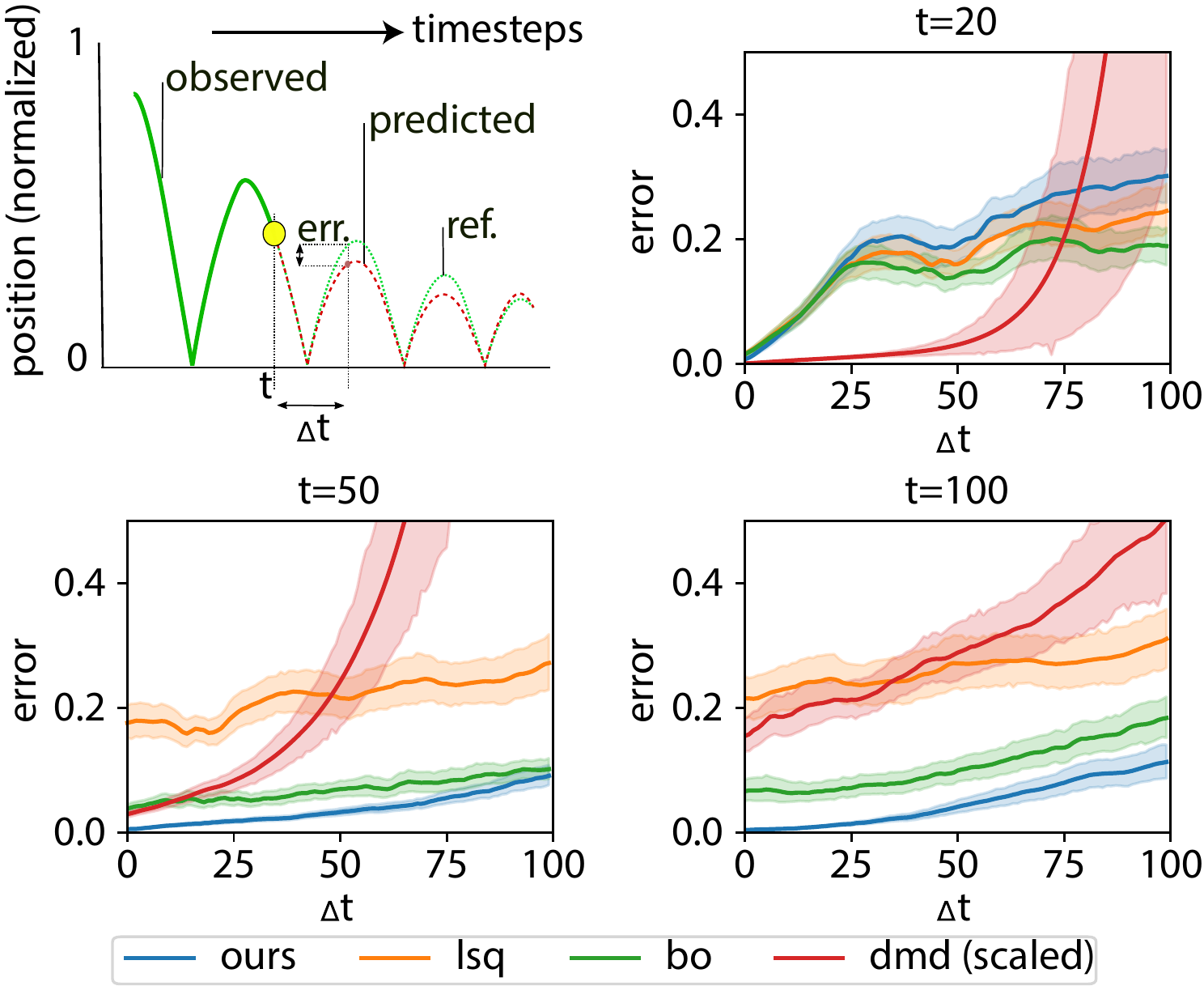}
\end{subfigure}%
\caption{\textbf{Accuracy of forward prediction.} The accuracy is evaluated after 20, 50 and 100 frames are observed, predicting for the next 100 frames. The DMD error is scaled down 1k, 50 and 5 times respectively for predictions after 20, 50 and 100 observations. We note \cite{wu2015galileo} as 'lsq', \cite{gonzalez2016gpyopt} as 'bo' and \cite{demo18pydmd} as 'dmd'.}
\label{forw_pred}
\vspace{\offset}
\end{figure}

\begin{figure}[htbp]
\centering
\includegraphics[width=0.46\textwidth]{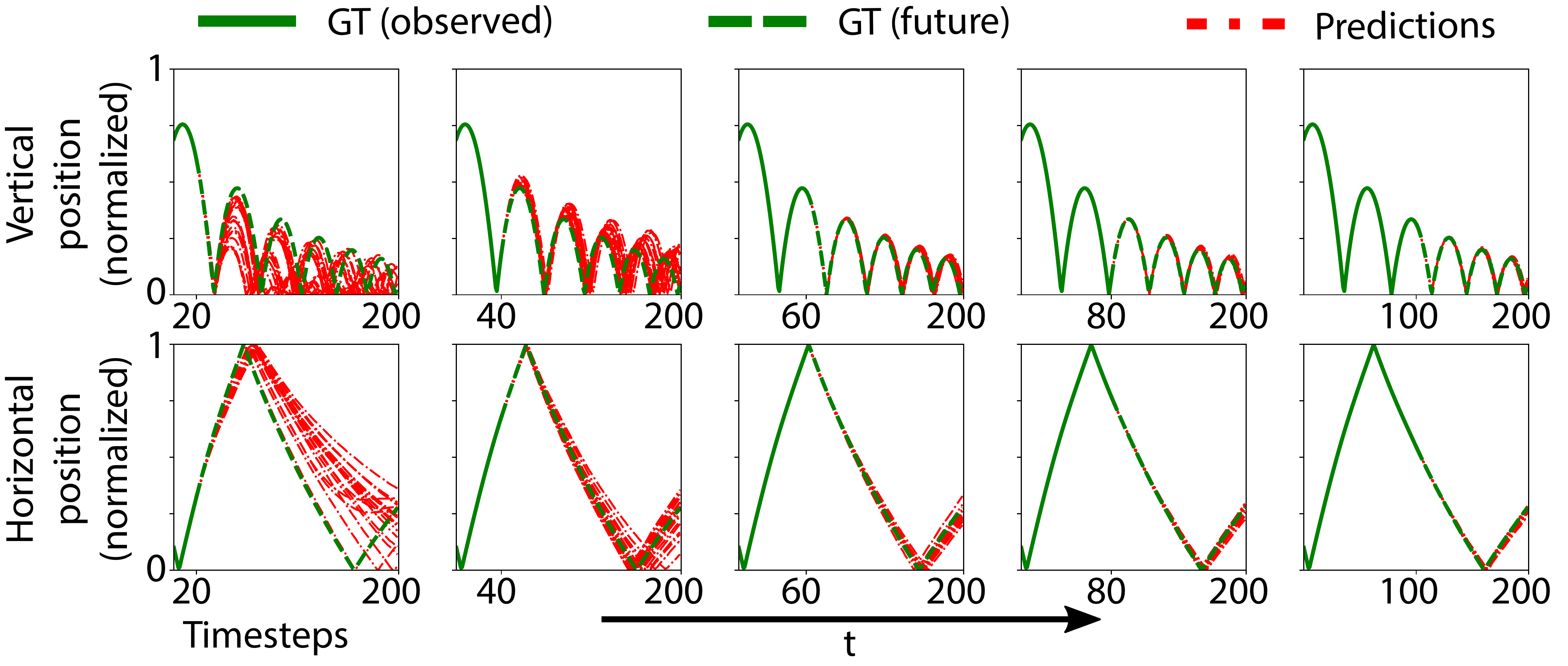}
\caption{\textbf{Forward prediction uncertainty} after different numbers of observed frames (20, 40, 60, 80 and 100). Observed trajectory (green), future ground truth (dashed-green), predictions (red).}
\label{forw_uncert}
\vspace{\offset}
\end{figure}

Here, we evaluate the future prediction accuracy as frames are observed. In
addition to previous baselines in Sec.\ref{SysIDexp}, we add an additional
non-parametric model for system identification that approximates responses using a set of modes with different frequencies  \cite{kutz2016dynamic}. Three
sets of predictions are evaluated - after 20, 50 and 100 frames respectively - until the end of the video at 200 frames, as shown in fig.\ref{forw_pred}. We visualize example model predictions and their associated uncertainty in fig.\ref{forw_uncert}.  Importantly, the proposed approach becomes more certain as additional frames are observed, highlighting the probabilistic nature of Vid2Param.

\subsection{Varying physical properties and sensitivity analysis}

\begin{figure}[htbp]
\centering
\includegraphics[width=0.46\textwidth]{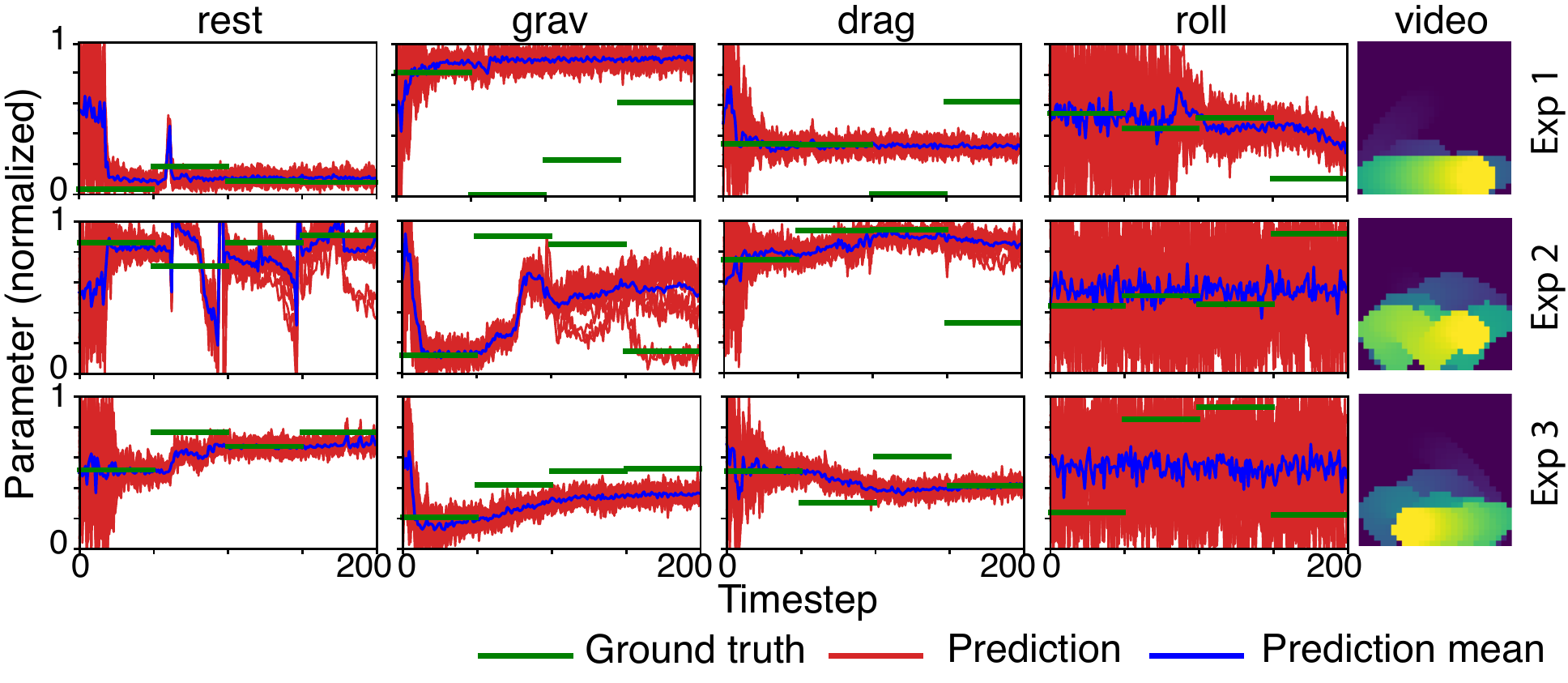}
\caption{\textbf{System Identification from video with varying physical parameters.} The physical parameters of the bouncing ball change every 50 frames (4 times per video). We plot ground truth (green), predicted samples (red), and the predicted - mean (blue). Given enough excitation, our model can detect the change in the parameters.}
\label{pics:var_physics}
\vspace{\offset}
\end{figure}

In this experiment, we evaluate how well Vid2Param can estimate physical
parameters when they are changing as the video is unrolled (using a model where parameters are assumed to stay constant throughout the video).
Therefore, this is a test of robustness or sensitivity of the model. We generate a new dataset, wherein the parameters change every 50 frames (2.5 seconds) in a given video. The results are shown in fig.\ref{pics:var_physics}, and show that the proposed model can infer changing parameters, provided there is enough system excitation to facilitate this. For example, gravitation coefficients can only be inferred if the ball is bouncing, while the rolling coefficient can be inferred if the ball is rolling.

\begin{figure}[htbp]
\centering
\includegraphics[width=0.49\textwidth]{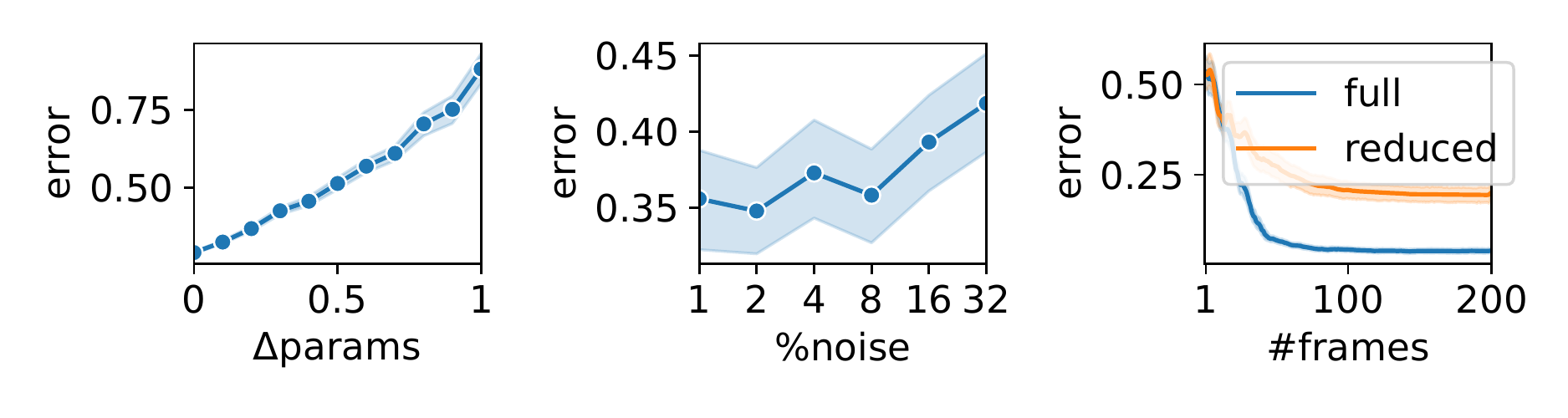}
\caption{\textbf{Sensitivity analysis.} (Left) Extrapolation of parameters
  (Middle) Training with noisy data (Right) Training with less accurate model - fixed drag and rolling coefficient.}
\label{pics:sensitivity}
\vspace{\offset}
\end{figure}

We also perform sensitivity analysis over multiple conditions. First, we test for extrapolation over unseen parameters by training on the first half of the range for different parameters - eg. $e_{train}\in[0.6,0.8]$ (for restitution).
We evaluate performance for increasing parameter deviations, $\Delta params$, in the test data - $e_{test}\in[0.8,1.0]$. Secondly, we
evaluate the performance when training with increasing additive parameter noise. Finally, we train  with
simpler physics model (fixed drag and rolling coefficient) and test how well we can estimate gravity and restitution when testing with the full model as the video is unrolled. Results can be seen in fig.\ref{pics:sensitivity}, with extrapolation performance slowly decaying outside of training regions and consistent performance even with large percentage of noise added to the training parameters.
\subsection{Real videos}

\begin{figure}[htbp]
\centering
\begin{subfigure}[c]{0.66\linewidth}
  \centering
  \includegraphics[width=1.0\linewidth]{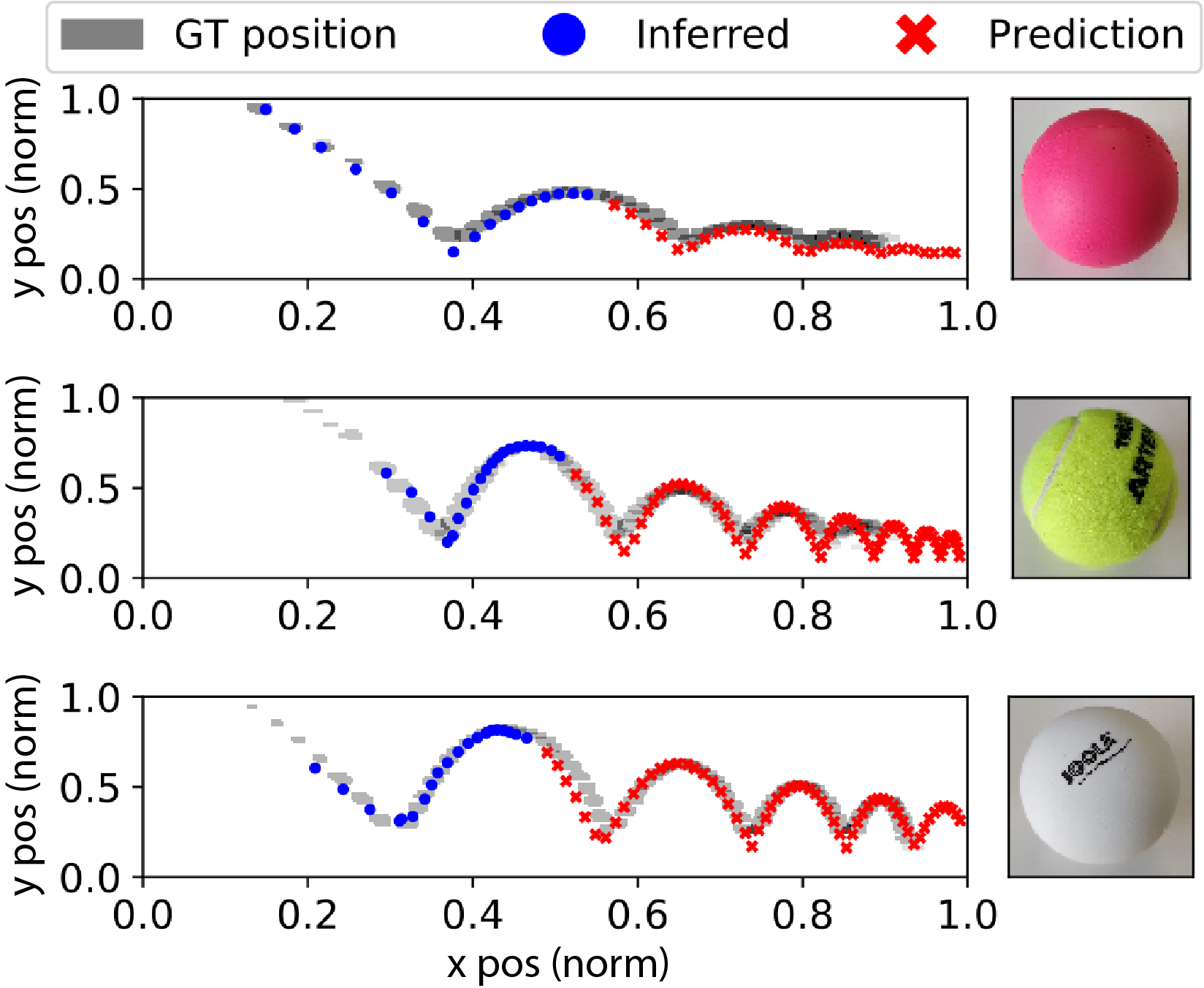}
  \caption{Convergence}
\end{subfigure}%
\begin{subfigure}[c]{0.33\linewidth}
  \centering
  \includegraphics[width=1.0\linewidth]{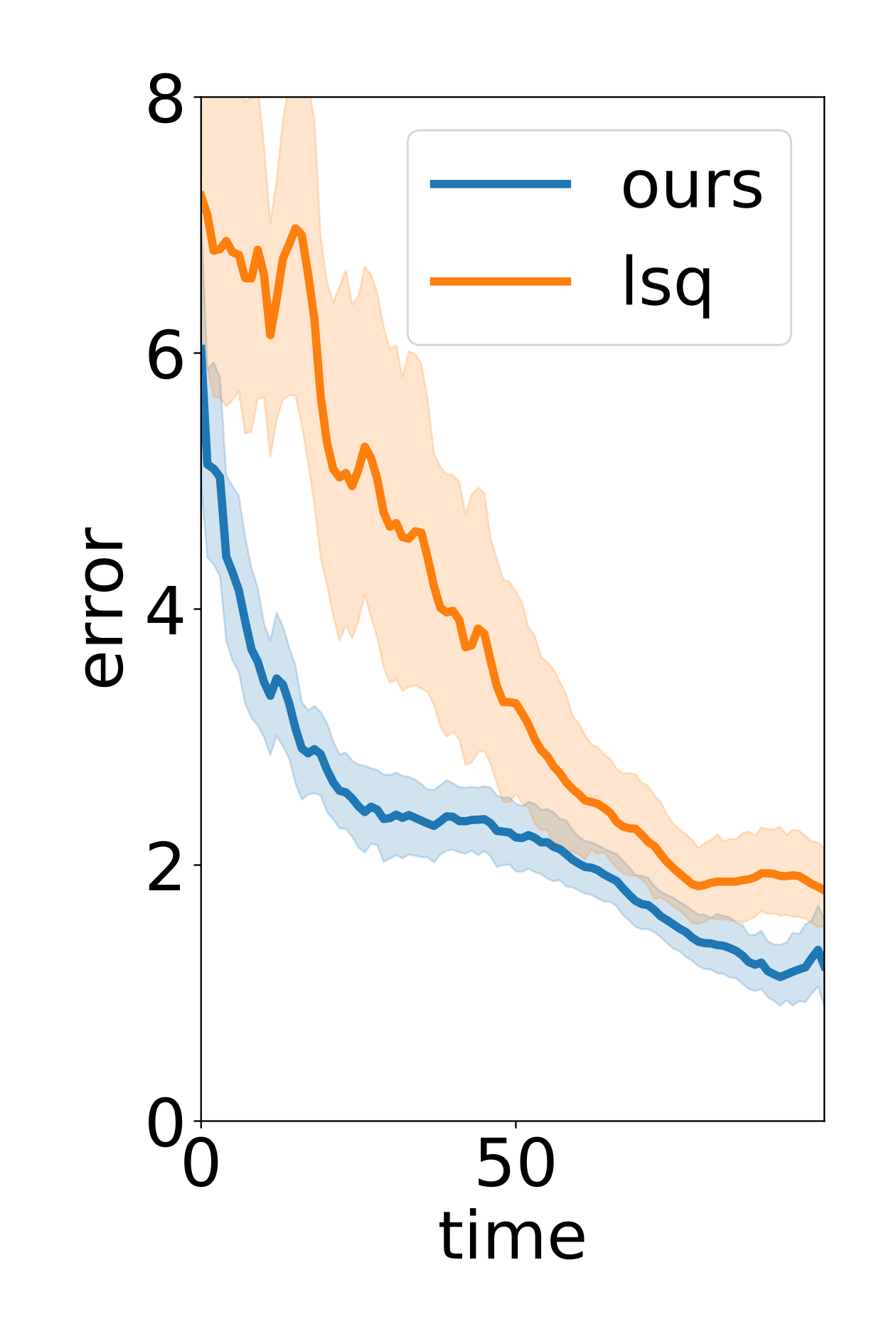}
  \caption{Accuracy}
\end{subfigure}
\caption{\textbf{Experiments with real videos.} We evaluate the performance of our method on real videos with bouncing balls with different physical properties (rubber, tennis, ping-pong). As the video is unrolled we compare the future predictions for the very last 10 frames of each video (the long term prediction accuracy) as explicit ground truth over the physical parameters is not available (a) Example model predictions for the three different types of balls, overlaid on the extracted positions from the images (b) Accuracy of the last 10 forward predictions as the video is observed. We denote \cite{wu2015galileo} as 'lsq'.}
  \label{realvideos}
\vspace{\offset}
\end{figure}

Here, we evaluate how well our method can scale to real videos. We record a set of videos, lasting between 1-5 seconds, of different types of bouncing balls - rubber, tennis and ping-pong balls. Since the exact physical properties of the balls are not available, we instead use the accuracy of the forward predictions as an evaluation metric. We compare the last ten positions of the ground truth position of the ball, with the forward predictions of a model as the video is unrolled. Here, we compare our method against \cite{wu2015galileo}, where we generate 5000 uniform samples of all physical parameters, horizontal and vertical velocities, horizontal and vertical position in a small region around the starting location of the ball - in order to account for some of the noise in the real videos. In fig.\ref{realvideos} we show the convergence of our method for different types of balls, as well as the accuracy of the forward predictions as the video is observed.

\begin{figure*}[htbp]
\vspace{2mm}
\begin{center}
\begin{minipage}[t]{.65\linewidth}
\vspace{0pt}
\centering
\includegraphics[width=1.0\linewidth]{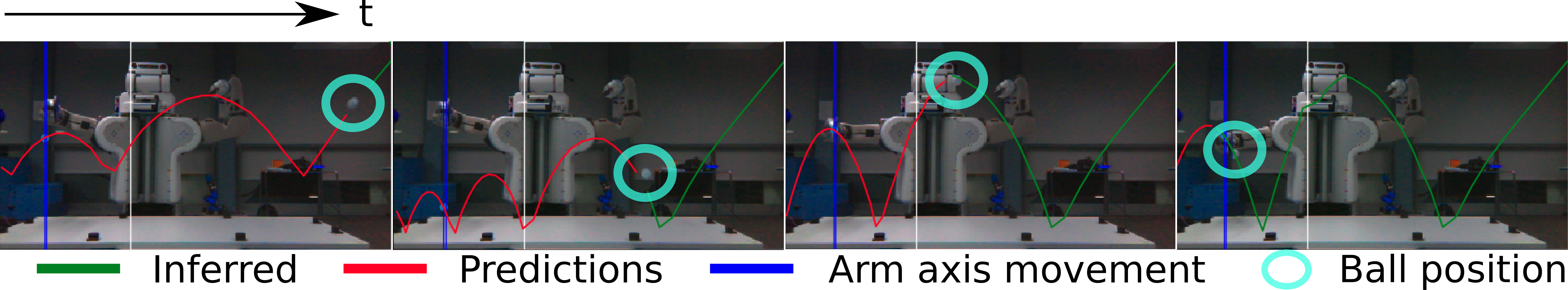}
\end{minipage}%
\begin{minipage}[t]{.34\linewidth}
\vspace{14pt}
\centering
\begin{tabular}{c|c} 
      \textbf{Random Policy} & 8/35 (23\%) \\
      \textbf{Random Policy (2x)} & 10/35 (29\%) \\
      \textbf{Vid2Param} & \textbf{27/35 (77\%)}\\
\end{tabular}
\end{minipage}
\end{center} 
\vspace{-4mm}
\caption{\textbf{Experiments with PR2.} 
 We perform a set of experiments, where a slowly actuated arm of the PR2 robot must intercept a bouncing ball, using only video as sensory input. We evaluate our method against a random policy feeding actions at 1Hz/2Hz. We demonstrated that our method can perform fast inference over the physical parameters of interest (e.g. restitution factor, height of the table) and correct future trajectories. Each experiment lasts about 2 seconds. At each throw of the ball, the model state is reset - so system identification is performed at every throw of the ball. (Please see the supplementary video for additional results.)}
\label{robotexpfig}
\vspace{\offsetl}
\end{figure*}
 \subsection{PR2 Robot experiments}

 Finally, we evaluate the accuracy and speed of our method in an experiment where the PR2 robot uses its arm to intercept a bouncing ball from a visual feed, using a standard low-cost camera as sensory input
 (please refer to supplementary video). Firstly, the camera is calibrated with
 respect to the arm movements, so that predictions of the ball in the image,
 correspond to the same position of the gripper. No calibration with respect to the bouncing surface, position/velocity mappings, size of the ball, etc. are needed since these should be robustly dealt with by the model trained on randomized physics in simulation.
 The difference between two consecutive
 frames are fed directly into our model and the latent predictions are unrolled
 until the future predicted horizontal position is approximately the same as the
 horizontal position of the gripper of the arm. Then the generated vertical
 position of the ball is sent as a positional set point to the arm. An experimental run
 usually lasts for 2-3 seconds, during which the PR2 robot must infer the physics of the
 ball, predict its future trajectory and execute an action to intercept it. Our
 model runs at $20$Hz on a standard laptop GPU, using IKFast for inverse kinematics of the arm. We use different types of ping-pong balls, in order to test how well our model can reach to balls with different physical properties. After each experiment (each throw of a ball), the model state is reset - so in each experiment we evaluate how well we can perform online system identification. Results can be seen in fig.\ref{robotexpfig}.

\section{Results and Analysis}

\textbf{Unified model for physical reasoning.}
In this work we present a model for inference
of physical parameters and generation of plausible future states from videos. We observe that such a model can be trained in an end-to-end fashion and perform accurate system identification in simulated and real settings, and subsequently used for control. We constrain our experiments to a single object and show that using just a video stream we can perform \textit{online} system identification. Importantly, the proposed approach is able to generalise well to images captured from a real camera, despite only being trained on simulated data. This highlights the value of sim2real techniques for interpreting physical parameters in various applications, and its potential to enable reasoning about physical properties, from relatively low fidelity sensors bootstrapped by learning from simulation.

\textbf{System identification.}
We observe that the proposed method can accurately infer different physical
parameters, outperforming baselines from the literature.
The magnitude of gravity and air drag can usually be inferred from observing just a few frames. Air drag can usually be inferred after observing a few more frames, as it is a function of both horizontal and vertical velocity, rather than just the vertical velocity as in the case of gravity. Restitution factors can be inferred a few frames after the ball has bounced for the first time. The rolling coefficient
has a higher error, which starts to decrease towards the end of the videos. While traditional system identification methods can infer physical properties,
which have a clear effect on the trajectory (such as restitution), it is
challenging to infer properties that jointly contribute to a certain effect. By
using domain randomization and a sim2real approach, our method can learn the
difference in parameters of trajectories with similar appearance even when trained
with noisy data. As such we can accurately estimate parameters with similar effect on the dynamics (such as gravity and air drag), as well as parameters whose effects are not
observed until the end of the trajectory (such as rolling coefficients).  Moreover, we
also demonstrate that to an extent detect change in the parameters, as a
video is unrolled (although this requires system excitation), and extrapolate to unseen parameters. 

\textbf{Forward predictions.} We have shown that our model can perform forward predictions in the latent space, over parameters of interest such as physical state variables. The forward predictions bring out key aspects of the evolution of uncertainty, such as high variability before a bounce and lower variability soon after, high variability over the stopping point before rolling is observed, etc. The proposed approach outperforms both parametric and non-parametric baselines in its ability to accurately perform forward predictions.

\textbf{Limitations and future work.} We observe in robotics experiments that our model performs well in real settings. Nevertheless, we experienced some limitations arising from making the predictions based on a single image, e.g. the ball passing behind or in front of the gripper. Thus in the future it will be beneficial to extend this line of work by inferring future predictions from multiple sources of video stream from different locations, or incorporating depth sensing. We note that our proposed model is independent of the choice of encoder, decoder and training data. It would be of interest to explore how such a model would perform with richer training data (e.g. multiple objects) by using advanced domain randomization \cite{zheng2018t2net} or different encoder-decoder structure \cite{battaglia2016interaction}.


\section{Conclusions}

This paper presents a method for online system identification from video. We
benchmark the proposed approach against existing baselines from the literature,
showing it outperforms these both in terms of speed and accuracy of
identification. We then demonstrate the utility of this approach with the task
of stopping a bouncing ball with a robot arm, performing online identification
from a camera feed and using the proposed model for inference of the physics
parameters. Further, we show its ability to generate future predictions of the
ball position, laying the groundwork for much more sophisticated predictive
motion planning schemes.





\bibliographystyle{abbrv}
\bibliography{bibl}  

\end{document}